\documentclass{article}
\usepackage{spconf,amsmath,epsfig}

\usepackage[]{graphicx}
\usepackage{caption}
\usepackage{subfig} 
\usepackage{amsmath}
\usepackage{amssymb}
\usepackage{epsfig}
\usepackage{cite}
\usepackage{color}
\usepackage{balance}
\usepackage{amsmath}
\usepackage{amsfonts} 
\usepackage{graphicx}
\usepackage{pdfpages}
\usepackage[T1]{fontenc} 
\usepackage{array}
\usepackage{url}

\usepackage{color}
\usepackage{wrapfig}
\usepackage{lipsum}
\usepackage[hidelinks]{hyperref}
\usepackage{multirow}
\usepackage{tabularx}
\usepackage{lipsum}
\usepackage{algpseudocode} 
\usepackage{algorithm,algpseudocode}

\begin{document}

\title{Detecting Rainfall Onset Using Sky Images}

\name{Soumyabrata Dev$^{1}$, Shilpa Manandhar$^{1}$, Yee Hui Lee$^{1}$, Stefan Winkler$^{2}$\thanks{This research is funded by the Defence Science and Technology Agency (DSTA), Singapore.}\thanks{Send correspondence to \url{Stefan.Winkler@adsc.com.sg.}}}
\address{
	$^{1}$~School of Electrical and Electronic Engineering, Nanyang Technological University (NTU), Singapore \\  
	$^{2}\,$Advanced Digital Sciences Center (ADSC), University of Illinois at Urbana-Champaign, Singapore \\
}

\maketitle

\begin{abstract}
Ground-based sky cameras (popularly known as Whole Sky Imagers) are increasingly used now-a-days for continuous monitoring of the atmosphere. These imagers have higher temporal and spatial resolutions compared to conventional satellite images. In this paper, we use ground-based sky cameras to detect the onset of rainfall. These images contain additional information about cloud coverage and movement and are therefore useful for accurate rainfall nowcast. We validate our results using rain gauge measurement recordings and achieve an accuracy of $89$\% for correct detection of rainfall onset.  
\end{abstract}

\section{Introduction}
Remote sensing analysts have traditionally used weather satellite images for predicting weather conditions. However, with the advent of low-cost photogrammetric techniques, there has been a paradigm shift in the study and analysis of several earth's events. Now-a-days, Whole Sky Imagers (WSIs) are extensively used for continuous monitoring of the earth's atmosphere~\cite{GRSM2016}. These imagers capture the sky at regular intervals of time and provide images of the earth's atmosphere with high temporal \emph{and} spatial resolution. 

In our research, we use ground-based WSIs to study the effects of clouds on satellite communication links. We compute cloud coverage and classify them into several cloud genera. We also use multiple sky cameras for 3D  reconstruction, base height estimation, and motion prediction of the clouds. In addition to imagers, we also use ground-based meteorological sensors for continuous monitoring of temperature, humidity, precipitation, etc. These different sensors help us in analyzing cloud features and weather formations methodically.  

In this paper, we use WSIs to detect the onset of rainfall. We can use images captured by WSIs for accurate rainfall nowcast. We present a technique to detect precipitation using the luminance of the captured image.

Traditionally, rainfall onset is studied using long-term climate data. Rainfall measurements from weather stations are analyzed to detect hidden patterns, and subsequently predict the onset of precipitation~\cite{Farahmand2014}. Precipitation radar is also used to measure tropical rainfall with high accuracy~\cite{TRMM2002}. However, these studies are based on long-term rain forecasting. Recently, sky cameras have been used to estimate the clear sky intensity distribution~\cite{Nou2015}. They are also used for short-term solar energy forecasting~\cite{Chow15}. However, to the best of our knowledge, no prior work have used sky cameras to detect rainfall onset. 

This paper is organized as follows: we discuss our experimental setup in Section~\ref{sec:setup} and the impact of precipitation in Section~\ref{sec:impact}. Our proposed methodology to detect precipitation and the results obtained are discussed in Section~\ref{sec:results}. Section~\ref{sec:conc} concludes the paper.

\section{Data Description}
\label{sec:setup}
Our measurements are recorded at several rooftops of Nanyang Technological University (NTU) Singapore. We continuously monitor the temperature, humidity, rainfall and solar radiation of the atmosphere using weather stations. We have also installed collocated WSIs that continuously capture sky images at regular time intervals. All these meteorological measurements and images are archived in a server.

\subsection{Whole Sky Imagers (WSI)}
\begin{figure}[htb]
\begin{center}
\includegraphics[width=0.25\textwidth]{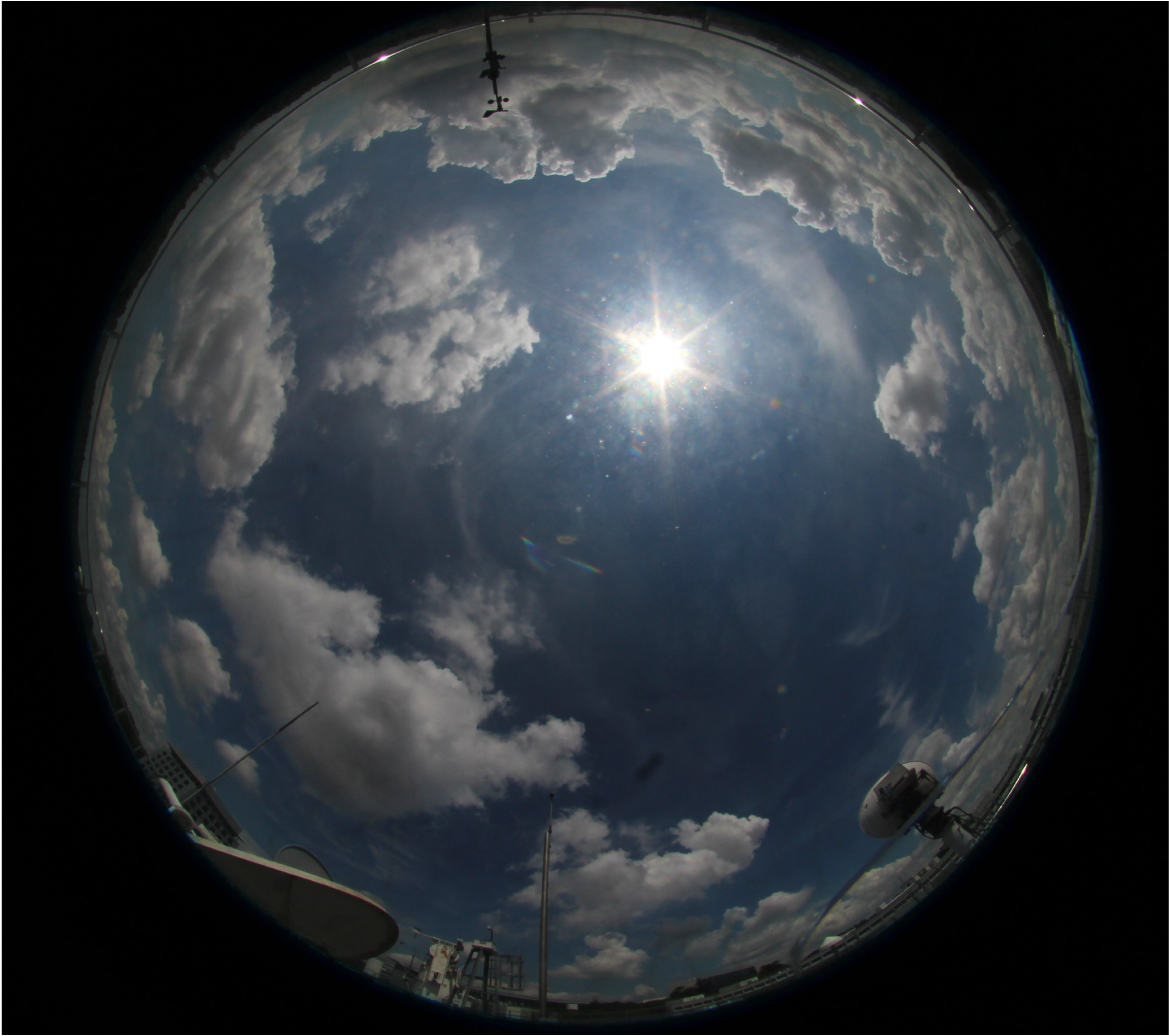}
\caption{Sample image captured by our sky camera at NTU. These imagers are collocated with weather stations.
\label{fig:all_WAHRSIS}}
\end{center}
\end{figure}

A Whole Sky Imager (WSI) is a ground-based, wide-angle sky camera that captures the sky scene at regular intervals of time. Figure~\ref{fig:all_WAHRSIS} shows an example image taken by our sky camera. We have designed and deployed several custom-built WSIs~\cite{WAHRSIS,IGARSS2015} that capture images at an interval of $2$ minutes. Currently, these imagers are deployed on the rooftops of our university buildings at $1.34^{\circ}$N, $103.68^{\circ}$E. 

\subsection{Weather Station}
Our weather stations are Davis Instruments $7440$ Weather Vantage Pro II. These are equipped with tipping buckets that measure the amount of rainfall in millimeters per hour. We use these rainfall recordings as ground-truth measurements in this paper. Additionally, the weather stations measure temperature, humidity and solar radiation. 

The solar radiation measured by the pyranometer that is part of the weather station typically follows a cosine response across different hours of the day.
Theoretically, the solar radiation measured at any given place on the earth's surface can be modeled using the clear-sky Global Horizontal Irradiance (GHI) model. We use the Singapore-specific version~\cite{dazhi2012estimation} given by:
\begin{align}
\label{eq:GHI-model}
G_c = 0.8277E_{0}I_{sc}(\cos\theta_z)^{1.3644}e^{-0.0013\times(90-\theta_z)},
\end{align}
where $G_c$ is the clear-sky GHI measured in W/$\mbox{m}^2$ and $\theta_z$ is the solar zenith angle (in degrees). $I_{sc}$ denotes the solar irradiance constant ($1366.1W/m^2$). The eccentricity correction factor of the earth $E_{0}$ is given by:
\begin{equation*}
\begin{aligned}
\label{eq:E0value}
E_0 = 1.00011 + 0.034221\cos(\Gamma) + 0.001280\sin(\Gamma) + \\0.000719\cos(2\Gamma) + 0.000077\sin(2\Gamma), 
\end{aligned}
\end{equation*}
where the day angle (measured in radians) is given by 
$\Gamma = 2\pi(d_n-1)/365$. The term $d_n$  indicates the day number in a year. We denote the solar radiation calculated using Eq.~\ref{eq:GHI-model} as \emph{clear-sky radiation}. 

\section{Impact of Precipitation}
\label{sec:impact}

In \cite{IGARSS_solar}, we observed that the luminance of the image in the circumsolar region (region around the sun) and the measured solar radiation from the weather station are correlated. Therefore, we can define the empirical luminance of the clear sky from the theoretical clear-sky model (cf.\ Eq.~\ref{eq:GHI-model}), based on the  relationship between solar irradiance and image luminance~\cite{IGARSS_solar}. We define this estimated luminance as the \emph{clear-sky luminance}, denoted by $L_c$. In other words, it measures the luminous intensity per second observed by our camera sensor, in the event of a clear sky.

In addition to clear-sky luminance, we also calculate the luminance of the whole sky image denoted by $L_m$. We crop a square of dimension $2000 \times 2000$ pixels from the image to exclude the occlusions and neighboring buildings. The average luminance of the cropped image is $L_m$.

\begin{figure*}[htb]
\centering
\includegraphics[width=0.8\textwidth]{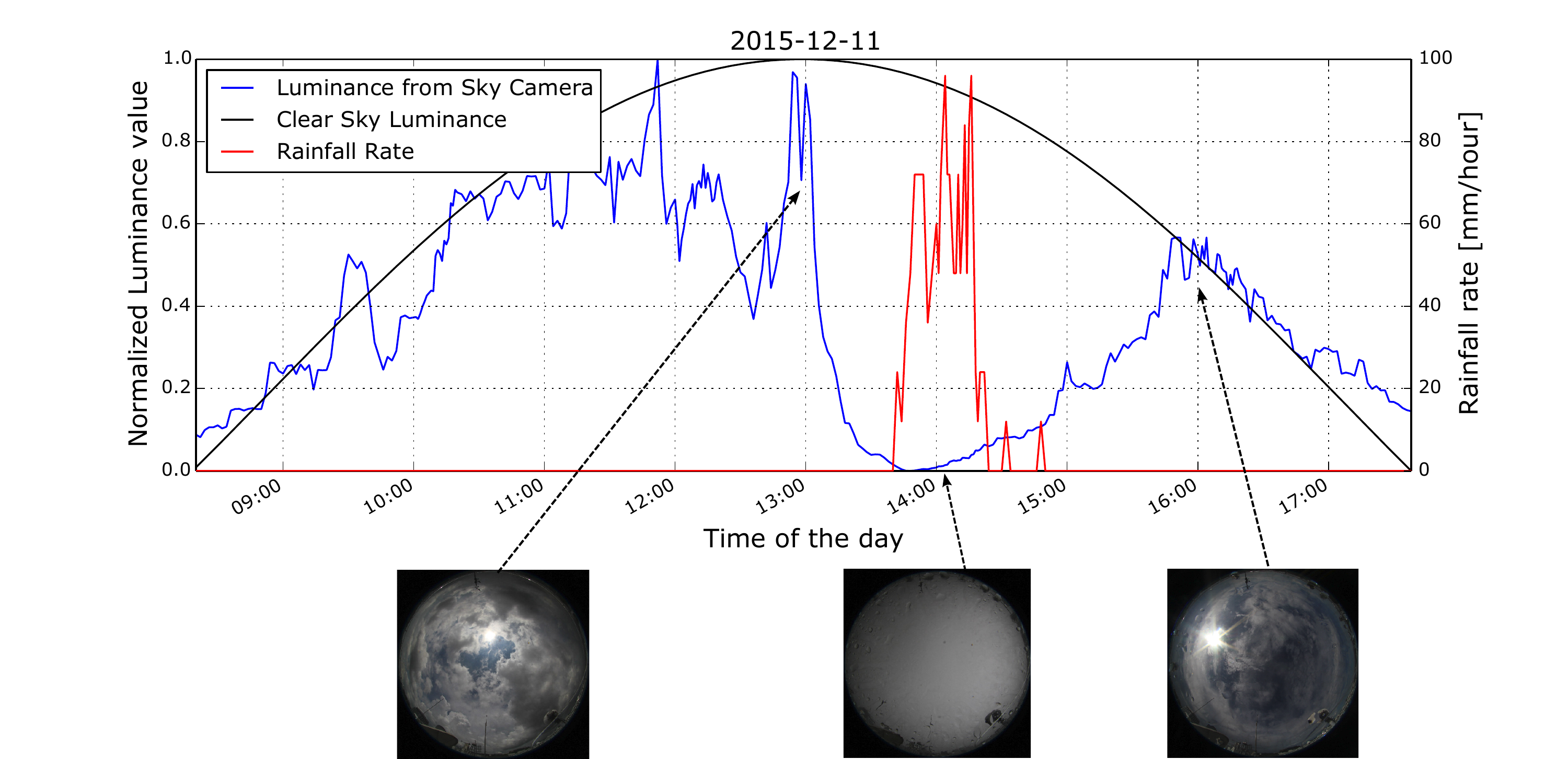}
\caption{We illustrate the measured luminance and clear-sky luminance for the $11^\mathrm{th}$ of December 2015. The primary y-axis shows the normalized luminance value, and the secondary y-axis measures the rainfall rate (measured in mm/hour). We also show the corresponding sky images captured by our sky camera at three distinct times -- before, during and after the rain event.} 
\label{fig:example}
\end{figure*}

Figure~\ref{fig:example} shows the luminance value of the images captured across different times of the day for a typical day in December. We also calculate the clear-sky luminance and plot it in Fig.~\ref{fig:example}.
We observe that the measured luminance value sharply decreases with the onset of rainfall. The luminance continues to remain low during the entire duration of the rainfall. It gradually increases after the precipitation event. We verify this by referring to the corresponding sky images captured by our WSI at different time instants. The sun is visible in the first image captured at 12:54 (before the rain event), when we record a normalized luminance of $0.7$. The second image captured at 14:07 (during the rain event) has dark clouds, with a much lower normalized luminance of $0.02$. Finally, the third image captured at 16:00 (after the rain event) shows a clearer sky condition, and the normalized luminance is $0.58$. Therefore, we observe that the onset of precipitation reduces the luminance measured in the sky images compared to the clear-sky luminance. 

\section{Methodology \& Results}
\label{sec:results}
In this section, we describe our methodology to detect precipitation from sky camera images. The measured luminance value deviates from the ideal luminance value (corresponding to clear sky) during a rainfall event. We use this deviation from clear sky luminance as a measure to indicate rainfall onset.

We define Clearness Luminance Index ($I$) as the ratio of measured luminance ($L_m$) value to the clear-sky luminance ($L_c$) value, $I = L_m/L_c$. 

\begin{figure}[htb]
\centering
\includegraphics[width=0.45\textwidth]{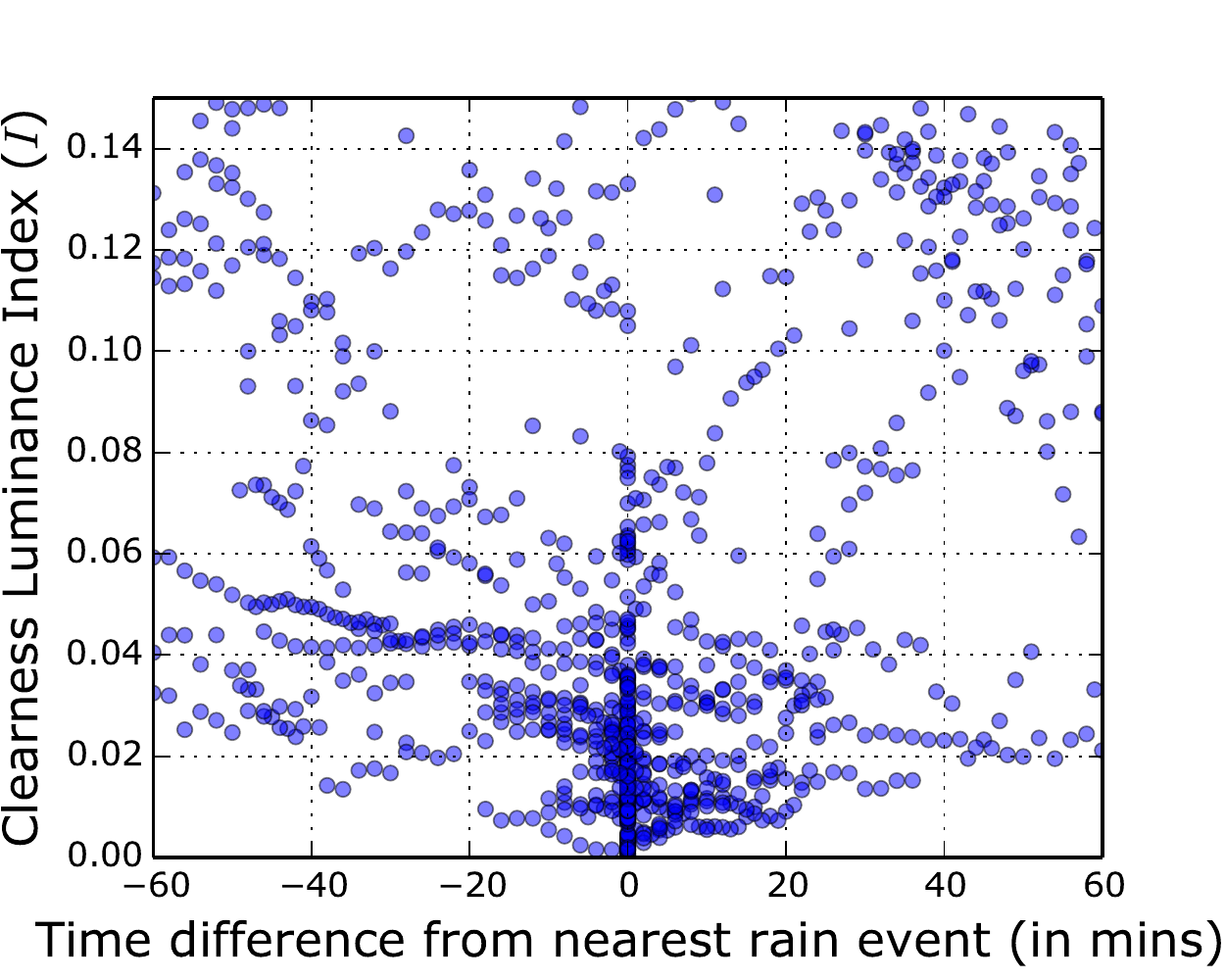}
\caption{Trend of Clearness Luminance Index ($I$) before and after precipitation measured for all days in December $2015$. The value of $I$ gradually decreases as it approaches rainfall onset, and then gradually increases again. } 
\label{fig:CLIvalues}
\end{figure}

In case of overcast condition, the obtained Clearness Luminance Index $I$ is low. On the other hand, $I$ equals unity during ideal clear-sky cases. We evaluate the values of $I$ for a  longer period of time, namely the entire month of December 2015. Figure~\ref{fig:CLIvalues} shows the relationship between index $I$ of the captured images and its corresponding distance from the nearest rain event (measured in minutes). 

From Fig.~\ref{fig:CLIvalues}, we observe that the index value gradually decreases as it approaches a rain event, and reaches a minimum during rainfall. Post rain event, the clearness luminance index gradually increases with the passage of time. The trend roughly follows a V-shaped curve.

\subsection{Critical Clearness Luminance Index}
In this section, we define a threshold in $I$ to detect the onset of rainfall as the Critical Clearness Luminance Index ($I_c$). The criterion for rainfall detection proposed here is as follows: if the measured index $I$ decreases below the critical $I_c$, there is an indication of precipitation. 

We consider all the images captured in December $2015$. We use rainfall recorded by the collocated weather station as ground truth measurements.
We classify each of the weather station measurements into \emph{rain} and \emph{non-rain} events, based on the rain gauge measurements. We also check the corresponding WSI images captured at the same time, and calculate their index $I$. The entire set of images in our dataset are segregated into two categories. The first category comprises those images that are within $\pm 15$ minutes of a rain event, including the images captured during the rain event. The rest of the images are considered in the second category.

\begin{figure}[htb]
\centering
\includegraphics[width=0.45\textwidth]{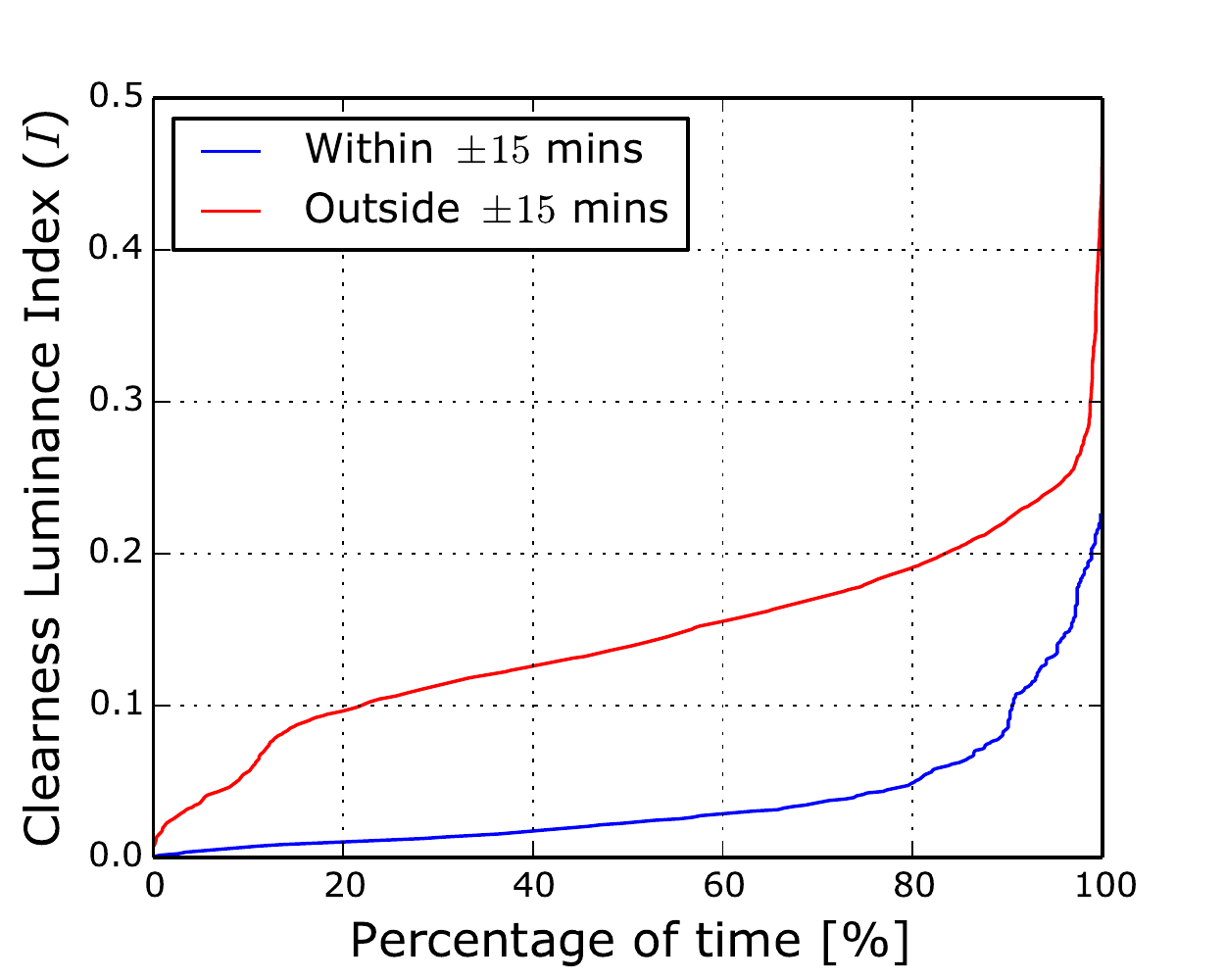}
\caption{CDF plots of Clearness Luminance Index of images within and outside $\pm 15$ minutes of rain event. Each point in the graph denotes the percentage of time the clearness index $I$ is less than its corresponding value.} 
\label{fig:cdf}
\end{figure}

Figure~\ref{fig:cdf} plots the Cumulative Distribution Plot (CDF) of clearness luminance index of images within and outside the time window of $\pm 15$ minutes. 

The motivation of classifying the images into two categories (within and outside $\pm 15$ minutes interval) is to analyze the impact of precipitation on the image luminance $L_m$.
From Fig.~\ref{fig:cdf}, we observe that images within $\pm 15$ minutes interval of rain event have a much lower luminance index than those images further from a rain event. This is intuitive as dark cumulus clouds appear before a rainfall, and they often persist for a brief period after the rainfall event. Furthermore, dark clouds are generally not present on non-rainy days.

\subsection{Operating Characteristics (OC) Curve}
\begin{figure}[htb]
\centering
\includegraphics[width=0.45\textwidth]{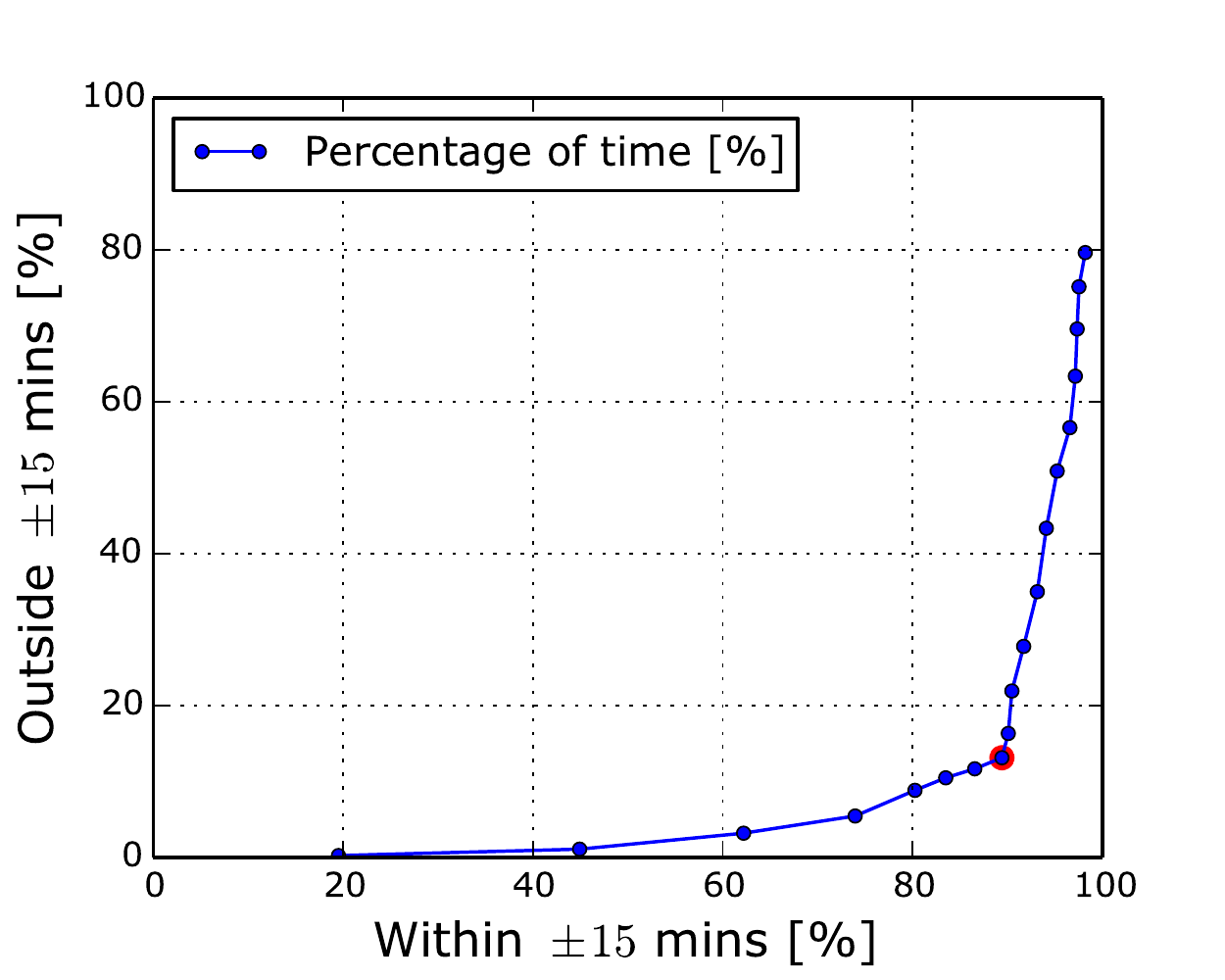}
\caption{Operating Characteristics (OC) curve generated by varying the critical clearness luminance index ($I_c$). The \emph{elbow} in the plot (shown in red) is our chosen $I_c$.} 
\label{fig:roc}
\end{figure}

In this section, we are interested in determining the critical clearness luminance index ($I_c$). For this purpose, we calculate the operating characteristics (OC), which shows the percentage of time for images within $\pm 15$ minutes of rain event against those  outside $\pm 15$ minutes of rain event. We vary the value of $I_c$ from $0.01$ to $0.2$, in steps of $0.01$. The resulting OC curve is shown in Fig.~\ref{fig:roc}.

We observe a sharp transition at $I_c = 0.08$. This is generally referred as the \emph{elbow} of the OC curve, and is chosen as the operating point as it has the best performance. 
We check the corresponding percentage of time for this critical threshold ($I_c$) of $0.08$. We observe that the images within $\pm 15$ minutes of rain events have a clearness index below the critical threshold for $89.41\%$ of time. On the other hand, only $13.13\%$ of time images outside the time window have a clearness index below the critical threshold. 
Therefore, $I_c=0.08$ can be used as the threshold for the clearness index for estimating the onset of rainfall. However, this threshold is derived for a tropical region like Singapore. This value may need to be adjusted when applying this model to other regions or using different sky imagers.

\section{Conclusion}
\label{sec:conc}
In this paper, we have used ground-based sky cameras to detect rainfall onset. We defined a Clearness Luminance Index that measures the deviation of estimated luminance from the clear-sky luminance. We have proposed a critical clearness luminance index that indicates the onset of precipitation. Extensive results on a month of sky images and weather station recordings show the efficacy of our proposed approach. In the future, we plan to verify these results over longer periods.

\bibliographystyle{IEEEbib}

\end{document}